\documentclass[conference]{IEEEtran}
\IEEEoverridecommandlockouts
\usepackage{fancyhdr}
\usepackage{cite}
\usepackage{amsmath,amssymb,amsfonts}
\usepackage{algorithm}
\usepackage{algorithmic}
\usepackage{graphicx}
\usepackage{textcomp}
\usepackage{xcolor}
\def\BibTeX{{\rm B\kern-.05em{\sc i\kern-.025em b}\kern-.08em
    T\kern-.1667em\lower.7ex\hbox{E}\kern-.125emX}}

\usepackage{microtype}
\usepackage{subfigure}
\usepackage{booktabs} % for professional tables
\usepackage{enumitem}
\setlist[itemize]{noitemsep, topsep=0pt, parsep=0pt, partopsep=0pt}
\usepackage{bm}
\usepackage{makecell} % Add this to your preamble for proper multi-line cells.
\usepackage{bbold}
\usepackage{hyperref}

\usepackage{mathtools}
\mathtoolsset{showonlyrefs}
\usepackage{amsthm}

\theoremstyle{plain}
\newtheorem{theorem}{Theorem}[section]

\theoremstyle{definition}

\newtheorem{assumption}[theorem]{Assumption}
\theoremstyle{remark}

\begin{document}

\title{Monte Carlo--Type Neural Operator for Differential Equations %\\
% {\footnotesize \textsuperscript{*}Note: Sub-titles are not captured for https://ieeexplore.ieee.org  and
}

\author{\IEEEauthorblockN{1\textsuperscript{st} Salah Eddine Choutri}
\IEEEauthorblockA{\textit{NYUAD Research Institute} \\
\textit{New York University Abu Dhabi}\\
Abu Dhabi, UAE\\
choutri@nyu.edu}
\and
\IEEEauthorblockN{2\textsuperscript{nd} Prajwal Chauhan}
\IEEEauthorblockA{\textit{Engineering Division} \\
\textit{New York University Abu Dhabi}\\
Abu Dhabi, UAE\\
pc3377@nyu.edu}
\and
\IEEEauthorblockN{3\textsuperscript{rd} Othmane Mazhar}
\IEEEauthorblockA{\textit{Laboratoire de Probabilités, Statistique et Modélisation} \\
\textit{Sorbonne University \& Université Paris Cité}\\
Paris, France \\
omazhar@lpsm.paris}
\and
\IEEEauthorblockN{\quad\quad  4\textsuperscript{th} Saif Eddin Jabari}
\IEEEauthorblockA{\quad\quad \textit{Engineering Division} \\
\quad\quad\quad \ \textit{New York University Abu Dhabi}\\ \quad\quad\quad
Abu Dhabi, UAE\\ \quad\quad \ 
sej7@nyu.edu}
}
\maketitle
\thispagestyle{fancy}
\begin{abstract}
The Monte Carlo-type Neural Operator (MCNO) introduces a framework for learning solution operators of one-dimensional partial differential equations (PDEs) by directly learning the kernel function and approximating the associated integral operator using a Monte Carlo-type approach. Unlike Fourier Neural Operators (FNOs), which rely on spectral representations and assume translation-invariant kernels, MCNO makes no such assumptions.
% It learns the kernel as a matrix over sampled input–output pairs, which defines a kernel integral operator that we approximate via Monte Carlo-type integration.
The kernel is represented as a
learnable tensor acting on the latent features at sampled points, with sampling performed once uniformly at random from the discretized grid. This design enables generalization across multiple grid resolutions without relying on fixed global basis functions or repeated sampling during training, while an interpolation step maps between arbitrary input and output grids to further enhance flexibility. Experiments on standard 1D PDE benchmarks show that MCNO achieves competitive accuracy with efficient computational cost. We also provide a theoretical analysis proving that the Monte Carlo estimator yields a bounded bias and variance under mild regularity assumptions. This result holds in any spatial dimension, suggesting that MCNO may extend naturally beyond one-dimensional problems. More broadly, this work explores how Monte Carlo-type integration can be incorporated into neural operator frameworks for continuous-domain PDEs, providing a theoretically supported alternative to spectral methods (such as FNO) and to graph-based Monte Carlo approaches (such as the Graph Kernel Neural Operator, GNO).
\end{abstract}


\section{Introduction}

Neural networks have shown remarkable success in learning mappings between finite-dimensional vector spaces, particularly in image, language, and scientific modeling tasks. Extending this capability to mappings between infinite-dimensional function spaces has led to the development of neural operators, a class of models that approximate solution operators for partial differential equations (PDEs). These models aim to learn the operator that maps problem inputs, such as coefficients, boundary conditions, or forcing terms, to PDE solutions, enabling rapid inference after training and generalization across parameterized problem families.

One of the earliest operator learning architectures, the Deep Operator Network (DeepONet)~\cite{lu2021learning}, uses separate networks (branch and trunk) to learn nonlinear mappings from input functions to solution function evaluations. More recently, the Fourier Neural Operator (FNO)~\cite{li2021fourier} introduced a spectral convolution approach, learning Fourier multipliers and leveraging the Fast Fourier Transform (FFT) to efficiently approximate the solution operator of a PDE via an integral formulation. FNO assumes the underlying integral kernel to be translation invariant and typically operates on uniform grids with periodic boundary conditions, making it efficient. However, it can struggle to represent localized or non-periodic phenomena \cite{Prajwal}.

To improve spatial localization and multi-scale modeling, wavelet-based neural operators have been proposed. The Multiwavelet Transform Operator (MWT)~\cite{gupta2021multiwavelet} and the Wavelet Neural Operator (WNO)~\cite{tripura2022wavelet} apply compactly supported wavelets to enable hierarchical and localized operator learning. Another direction is the Convolutional Neural Operator (CNO)~\cite{raonic2023convolutional}, which adapts convolutional networks to operator learning by applying spatial filters over structured grid data. This design enables local feature modeling and performs well on problems with regular grid discretizations.

A related approach is the Graph Neural Operator (GNO)~\cite{li2020neuralb}, which extends operator learning to irregular spatial domains using graph message passing. GNO approximates the integral kernel through local aggregation across graph edges, resembling Monte Carlo integration in the graph setting. While it supports unstructured meshes, GNO has been evaluated on structured grids represented as regular graphs. In contrast, our suggested neural operator approximates the integral operator without relying on graph construction or repeated neighborhood aggregation, offering a simpler and more direct formulation.

In this work, we propose the Monte Carlo-type Neural Operator (MCNO), a lightweight architecture for efficient operator learning. MCNO learns the kernel of an integral operator directly and approximates it using a Monte Carlo-based approach with only a single random sample from the grid at the start of training. By avoiding spectral transforms and deep hierarchical architectures, it offers a simple alternative that balances computational efficiency and accuracy.
% We also provide theoretical bias and variance bounds under mild regularity assumptions.
While our benchmarks focus on PDEs posed on regular grids with periodic or homogeneous structure, MCNO itself does not assume translation invariance or rely on global basis representations. Its reliance on pointwise sampling, rather than domain-specific structure, suggests potential adaptability to unstructured or heterogeneous settings, although such cases are not explored in this work.

We evaluate MCNO on standard one-dimensional PDEs, including the Burger's and Korteweg–de Vries (KdV) equations. Within this setting, MCNO demonstrates competitive performance compared to established neural operator models, while maintaining architectural simplicity. We also provide a theoretical analysis showing that the Monte Carlo approximation yields a bounded bias and variance under mild regularity assumptions, and note that the result applies in arbitrary spatial dimension, suggesting the potential for broader applicability. \\

\textbf{Contributions.} The key contributions of this paper:
\begin{itemize}
    \item We introduce MCNO, a neural operator architecture that uses Monte Carlo-type integration with one-time random sampling to approximate solution operators for PDEs.
    \item We integrate feature-channel mixing directly into the Monte Carlo aggregation, enabling the kernel to capture both spatial and cross-channel dependencies.
    \item We incorporate an interpolation mechanism to reconstruct outputs on structured grids from sampled evaluations, enabling compatibility with standard data formats.
    \item We provide a theoretical bias bound for the Monte Carlo approximation under regularity assumptions.
    \item We demonstrate the effectiveness of MCNO on PDE benchmarks, showing competitive performance.
\end{itemize}

This paper is organized as follows: Section 2 presents the basics of neural operator learning. Section 3 introduces the MC Neural Operator. Section 4 outlines the experimental setup, datasets, training procedures and a comprehensive comparison with existing methods. Finally, Section 5 concludes with a discussion of potential extensions and future directions.

\color{black}

\section{Learning Neural Operators}
Neural operators are a robust framework for approximating mappings between infinite-dimensional function spaces, enabling efficient solutions to parametric partial differential equations (PDEs). This framework, originally proposed in works such as \cite{li2021fourier, kovachki2023neural, lu2021learning}, builds upon the idea of generalizing across resolutions, making neural operators flexible and computationally efficient. Unlike traditional numerical solvers, which are often tied to specific discretizations, neural operators leverage mesh-invariant architectures to enable tasks such as optimization, inverse modeling, and real-time simulations. These properties make neural operators particularly well-suited for scenarios requiring repeated PDE evaluations, such as those encountered in engineering and scientific applications.

\subsection{Operator Approximation}
The goal of neural operators is to learn a non-linear operator \( G^\dagger: A \to U \), where \( A = A(D; \mathbb{R}^{d_a}) \) and \( U = U(D; \mathbb{R}^{d_u}) \) represent Banach spaces of functions defined on a bounded domain \( D \subset \mathbb{R}^d \). From a dataset of input-output pairs \( \{a_j, u_j\}_{j=1}^N \), where \( a_j \) is sampled from a probability distribution \( \mu \) and \( u_j = G^\dagger(a_j) \), the aim is to construct an approximation:
\[
% G: A \times \Theta \to U \quad \text{or equivalently,} \quad
G_\theta: A \to U, \quad \theta \in \Theta,
\]
where \( \Theta \) is a finite-dimensional parameter space. The parameters \( \theta \) are optimized by minimizing a loss function that measures the discrepancy between predicted and actual outputs:
\[
\min_{\theta \in \Theta} \mathbb{E}_{a \sim \mu} \big[C(G(a, \theta), G^\dagger(a))\big].
\]

\subsection{Iterative Update Framework.}
Neural operators are structured as iterative architectures. The input function \( a \in A \) is first transformed into a higher-dimensional representation \( v_0(x) = P(a(x)) \) using a local map \( P \), typically implemented as a fully connected neural network. The representation is then updated through a sequence of transformations:
\[
v_{t+1}(x) := \sigma\big(Wv_t(x) + (\mathcal{K}(a; \phi)v_t)(x)\big), \quad \forall x \in D,
\]
where \( W: \mathbb{R}^{d_v} \to \mathbb{R}^{d_v} \) is a linear operator, \( \sigma \) is a non-linear activation function, and \( \mathcal{K}(a; \phi)v_t \) is a kernel integral operator parameterized by \( \phi \). After \( T \) iterations, the final representation \( v_T(x) \) is projected back to the target space \( U \) through another local map \( Q \).

\subsection{Kernel Integral Operator.}
The kernel integral operator defines the non-local updates in the iterative framework. It is expressed as: \small
\begin{equation} \label{kernel}
(\mathcal{K}(a; \phi)v_t)(x) := \int_D \kappa_\phi(x, y, a(x), a(y))v_t(y) \, dy, \ \ \forall x, % \in D,
\end{equation} \normalsize
where $\kappa_\phi$ is a learnable tensor kernel that may depend on the 
spatial locations, input features, and the local latent representation $v_t(y)$.

This formulation extends the concept of neural networks to infinite-dimensional spaces, enabling the approximation of highly non-linear operators through the composition of linear integral transformations and non-linear activation functions.

\section{MC-type Neural Operator Architecture}

\begin{figure}[h!]
    \centering
    \includegraphics[width=0.5\textwidth]{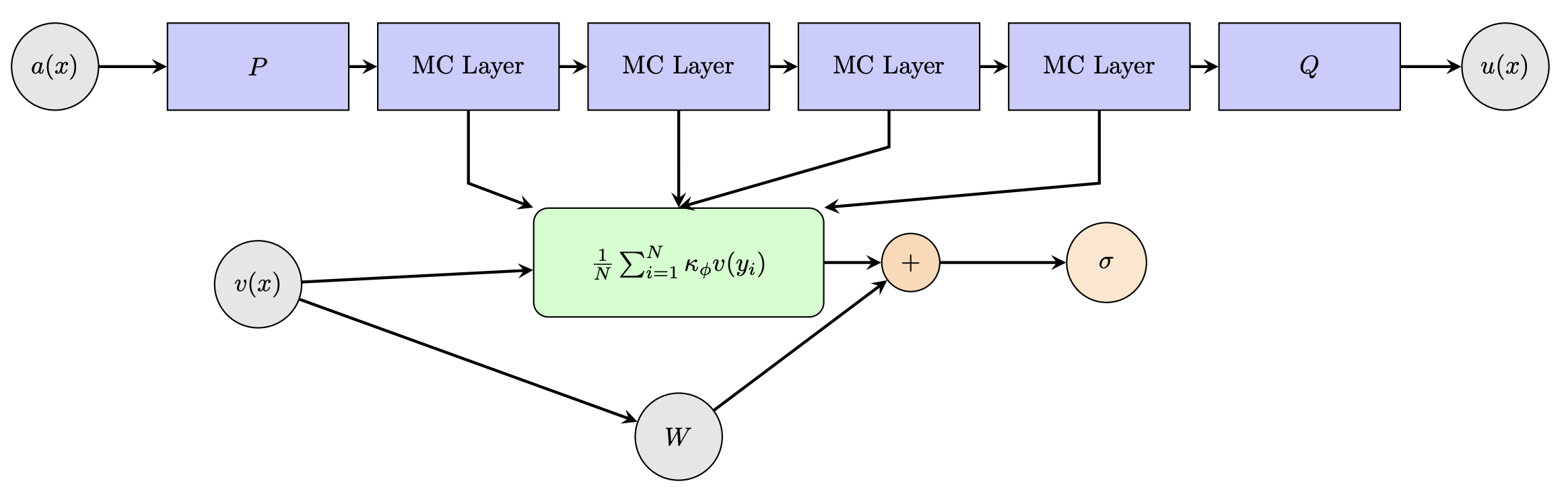} % Adjust width as needed
    \caption{MCNO's architecture} % Optional: Add a caption
    \label{fig:Picture3} % Optional: Add a label for referencing
\end{figure}

The MCNO architecture follows the same iterative framework described in Section 2. The input function \( a(x) \) is lifted to a higher-dimensional representation \( v_0(x) = P(a(x)) \), which is iteratively updated using the kernel integral operator. At each step, the update is given by:
\[
v_{t+1}(x) = \sigma \left( Wv_t(x) + (\hat{\mathcal{K}}_\text{MC}(a; \phi)v_t)(x) \right), \quad x \in D,
\]
where \( W \) is a linear transformation, \( \sigma \) is a non-linear activation function, and \( \hat{\mathcal{K}}_\text{MC}(a; \phi)v_t \) is the Monte Carlo-based kernel integral operator. From now on, for ease of notation, we will denote the kernel integral and its MC estimator by \( \mathcal{K}_\phi v_t\) and \( \hat{\mathcal{K}}_{N,\phi} v_t\) respectively. The general architecture of MCNO is illustrated in Figure \ref{fig:Picture3}.

\subsection{Monte Carlo Approximation for Integral Operators}

The Monte Carlo-type Neural Operator (MCNO) uses Monte Carlo integration to approximate the integral operator defined in~\eqref{kernel}. Instead of applying deterministic quadrature or spectral transforms, MCNO estimates the operator via a simple Monte Carlo average:
\[
\frac{1}{N} \sum_{i=1}^N \kappa_\phi(x, v_t(y_i)) v_t(y_i),
\]
where \( \{y_i\}_{i=1}^N \) are points sampled uniformly at random from a discretized computational grid, with sampling performed once at the start of training.
This fixed-sample strategy avoids repeated sampling during training, supports efficient batching, and integrates with structured-grid interpolation. %The method is fully differentiable and lightweight to implement.
\begin{algorithm}[h]
\caption{MCNO — Forward Pass}
\begin{algorithmic}[1]
\STATE \textbf{Given:} parameters $\phi$, $W$; samples $\{y_i\}_{i=1}^N$; grid; activation $\sigma$
\STATE \textbf{Input:} current representation $v_t(x)$
\STATE \textbf{Output:} $v_{t+1}(x)$
\STATE Estimate kernel: $$(\hat{\mathcal{K}}_{N,\phi}v_t)(x)=\tfrac{1}{N}\sum_{i=1}^N \kappa_\phi\!\big(x, v_t(y_i)\big)\, v_t(y_i)$$
\STATE Interpolate $(\hat{\mathcal{K}}_{N,\phi}v_t)$ onto the full grid
\STATE Update: $v_{t+1}(x)=\sigma\!\left(W v_t(x)+(\hat{\mathcal{K}}_{N,\phi}v_t)(x)\right)$
\end{algorithmic}
\end{algorithm}

\subsection{Efficient kernel interactions and feature mixing}

MCNO leverages Einstein summation (via PyTorch's einsum operation) to efficiently compute interactions between the learnable kernel tensors $\phi = \{\phi_i\}_{i=1}^N$ and the sampled latent features, i.e., 
$\kappa_\phi(x, v_t(y_i)) = \phi_i v_t(y_i)$, 
where each $\phi_i \in \mathbb{R}^{d_v \times d_v}$ is a learnable tensor associated with sample index~$i$.

Our implementation enables efficient batched computation and allows for GPU parallelization without requiring per-sample explicit matrix operations. Because each tensor $\phi_i$ acts directly on the feature channels, the kernel 
jointly aggregates information from the sampled spatial locations and mixes 
features across channels. 

Unlike architectures that separate spatial aggregation 
from channel mixing, MCNO performs both within a unified operation, increasing 
expressiveness without adding computational overhead.

\subsection{Interpolation for grid reconstruction}
Since the Monte Carlo integral is computed at a subset of sampled points, interpolation is required to reconstruct a structured grid representation compatible with standard neural architectures. This step maps sparse Monte Carlo estimates onto a dense discretized domain while preserving the local function structure.

MCNO employs continuous interpolation, typically linear for smooth functions, though 
nearest-neighbor interpolation can be used when sharper transitions must be preserved. Because the reconstruction involves only local averaging and scales linearly with the number of grid points $N_{\text{grid}}$, it contributes modestly to the overall cost compared to the kernel evaluation.

\subsection{Linear Complexity} 
MCNO achieves linear computational complexity with respect to the number of samples \(N\), offering an efficient method for approximating kernel integral operators. In our implementation, the kernel takes the form $\kappa_\phi(x, v_t(y_i))=\phi_i v_t(y_i)$, with each $\phi_i \in \mathbb{R}^{d_v \times d_v}$. This allows MCNO to compute kernel interactions only at the $N$ sampled points rather than across the full grid. This structure is a special case of the general kernel form used in our theoretical analysis.
In contrast, the Fourier Neural
Operator (FNO) computes the integral operator using the
Fast Fourier Transform (FFT), which scales quasi-linearly as
\(\mathcal{O}(N_{\text{grid}} \log N_{\text{grid}})\), where \(N_{\text{grid}}\) is the number of
discretization points. While efficient in practice, the FFT
introduces a logarithmic overhead that can become significant
on large grids or in high-dimensional settings.

The Monte Carlo estimator further admits theoretical error guarantees. Under mild regularity assumptions, we show that the bias decreases with the grid resolution $N_{\text{grid}}$, while the variance decays with the number of samples $N$.

\subsection{Bounding the Monte Carlo estimation error}
Consider the integral operator $\mathcal{K}_\phi$ defined as
\begin{equation}
\left( \mathcal{K}_\phi v_t \right)(x) = \int_D \kappa_\phi(x, y) v_t(y) \, dy,
\end{equation}
and the Monte Carlo estimator for the kernel integral operator
\begin{equation}
\left( \hat{\mathcal{K}}_{N, \phi} v_t \right)(x) = \frac{1}{N} \sum_{i=1}^N \kappa_\phi(x, y_{s_i}) v_t(y_{s_i}),
\end{equation}
where $D \subset \mathbb{R}^d$ is a compact domain. Let $\{y_j\}_{j=1}^{N_{\text{grid}}}$ be a uniform grid discretization of $D$ with spacing $h$ in each dimension, such that $N_{\text{grid}} = (1/h)^d$. 
\begin{assumption}\label{assum:RegultarityOfKv}
To proceed we make the following assumptions:
\begin{enumerate}
    \item Uniform Boundedness: There is $C > 0$ s.t $\forall \phi$, $x, y \in D$, and $v_t$:
    $\left| \kappa_\phi(x, y) v_t(y) \right| \leq C$.
    \item Uniform Lipschitz Continuity in $x$: There is $L_x > 0$ s.t $\forall \phi$, $y \in D$, and $x, x' \in D$:
    $\left| \kappa_\phi(x, y) v_t(y) - \kappa_\phi(x', y) v_t(y) \right| \leq L_x \|x - x'\|$, and $L > 0$ s.t $ \forall y, z \in D$:  $|\kappa_\phi(x, y) v_t(y) - \kappa_\phi(x, z) v_t(z)| \leq L \|y - z\|$.

% \item Boundedness: There is $C>0$ s.t $\forall x, y \in D$ $\phi$, $x$, $y$, and $v_t$ we have $|\kappa_\phi(x, y) v_t(y)| \leq C$.
\end{enumerate}    
\end{assumption}

\begin{theorem}\label{thm:MCEstimationError}
The Monte Carlo estimator $\hat{\mathcal{K}}_{N, \phi} v_t(x)$ approximates the integral operator $\left( \mathcal{K}_\phi v_t \right)(x)$ by averaging over $N$ points $\{y_{s_i}\}_{i=1}^N$ sampled uniformly at random from the grid and under the assumptions \ref{assum:RegultarityOfKv} we get with probability at least $1 - \delta$ the following estimation error bound
\begin{align}
&\sup_{x \in D} \left| \left( \hat{\mathcal{K}}_{N, \phi} v_t \right)(x) - \left( \mathcal{K}_\phi v_t \right)(x) \right| \\
&\qquad\leq C \cdot \frac{\sqrt{d}}{2} \cdot \text{Vol}(D) \cdot L \cdot N_{\text{grid}}^{-1/d} \\
&\qquad\qquad+\frac{3}{2} C \sqrt{\frac{2 \left( d \log N_{\text{grid}} + \log\left( \frac{2}{\delta} \right) \right)}{N}}.
\end{align}
\end{theorem}
This result ensures that the estimation error converges to zero with high probability for larger $N_{\text{grid}}$ and $N$. The proof of this result is given in the appendix and relies on a bias-variance decomposition and high probability concentration inequalities.
\section{Numerical experiments}

We evaluate the Monte Carlo-type Neural Operator on two standard benchmark problems: Burgers' equation and Korteweg-de Vries (KdV) Equation. These benchmarks are widely used for testing operator learning methods due to their diversity in  complexity and relevance to practical applications.
% Burgers' equation provides a testbed for one-dimensional nonlinear dynamics.
To ensure fairness and consistency, we adopt the same datasets and experimental setups as described in \cite{li2021fourier}.

MCNO is compared against several state-of-the-art operator learning methods and machine learning baselines. Table~\ref{tab:benchmarks} provides an overview of the methods, their description, and references.
\begin{table} [h]
\centering
\caption{Benchmark models evaluated in comparison to MCNO.}
\label{tab:benchmarks}
\resizebox{\columnwidth}{!}{%
\begin{tabular}{|l|p{3.5cm}|l|}
\hline
\textbf{Initials} & \textbf{Model Description}                                  & \textbf{Reference}           \\ \hline
FNO               & Fourier Neural Operator                                     & \cite{li2021fourier}          \\ \hline
WNO               & Wavelet Neural Operator                                     & \cite{tripura2022wavelet}    \\ \hline
MWT              & Multiwavelet Neural Operator                                 & \cite{gupta2021multiwavelet} \\ \hline
MGNO              & Multipole Graph Neural Operator                             & \cite{li2020multipolea}      \\ \hline
GNO               & Graph Neural Operator                                       & \cite{li2020neuralb}      \\ \hline
LNO               & Low-Rank Neural Operator, similar to DeepONet               & \cite{lu2021learning}          \\ \hline
% PCANN             & Principal Component Analysis Neural Network                 & \cite{bhattacharya2020model} \\ \hline
% RBM               & Reduced Basis Method                                        & \cite{devore2024reduced}     \\ \hline
% FCN               & Fully Connected Neural Network                              & \cite{zhu2018bayesian}       \\ \hline
% ResNet            & Residual Neural Network                                     & \cite{he2016deep}            \\ \hline
% % U-Net             & \makecell[l]{U-shaped Convolutional \\ Neural Network for Image Segmentation} & \cite{ronneberger2015unet} \\ \hline
% U-Net             & U-shaped Convolutional Neural Network for Image Segmentation & \cite{ronneberger2015unet} \\ \hline
% TF-Net            & Turbulent Flow Neural Network                               & \cite{wang2020towards}       \\ \hline
% FNO-2d            & 2D Fourier Neural Operator with RNN                         & \cite{li2021fourier}          \\ \hline
\end{tabular}%
}
\end{table}
The MCNO employs a four-layer architecture of Monte Carlo kernel integral operators, with ReLU activations. Training uses 1000 samples, with 100 additional samples reserved for testing. Optimization is performed with the Adam optimizer, starting at a learning rate of 0.001, halved every 100 epochs over a total of 500 epochs. We set $d_v=64, \ N=100$ for Burger's equation and $N=75$ for Korteweg-de Vries (KdV) equation. All experiments are conducted on an Tesla V100-SXM2-32GB of memory GPU and the accuracy is reported in terms of relative $L_2$ loss. Across both benchmarks, MCNO outperforms FNO and achieves competitive results in general, while maintaining linear computational complexity in the integration step.
% These results highlight MCNO's robustness and scalability as a framework for operator learning. 
The  errors for the main benchmarks FNO, MWT and WNO are reported with our implementation on the same GPU that we used to train our model. For the other benchmarks: GNO, LN and MGNO, the errors are taken from \cite{gupta2021multiwavelet}. In this work the models were trained and tested on an Nvidia V100-32GB GPU, therefore the comparison with their reported errors is fair to a certain extent.

\subsection{Burgers' Equation}

The one-dimensional Burgers' equation is a nonlinear PDE commonly used to model viscous fluid flow. The equation is expressed as:
\begin{align} \notag
   \partial_t u(x, t) &+ \partial_x \left( \frac{u^2(x, t)}{2} \right) = \nu \partial_{xx} u(x, t), \\
    &x \in (0, 1), \, t \in (0, 1], \notag
\end{align}
% \partial_t u(x, t) + \partial_x \left( \frac{u^2(x, t)}{2} \right) = \nu \partial_{xx} u(x, t), \\ x \in (0, 1), \, t \in (0, 1],
% \]
with periodic boundary conditions and initial condition \( u(x, 0) = u_0(x) \). Here, \( u_0 \) lies in \( L^2_{\text{per}}((0, 1); \mathbb{R}) \), and \( \nu > 0 \) is the viscosity coefficient.

\subsubsection{Dataset and Problem Setup}
The initial condition \( u_0(x) \) is generated from a Gaussian distribution \( \mu = \mathcal{N}(0, 625(-\Delta + 25I)^{-2}) \), where periodic boundary conditions are applied. The equation is solved using a split-step method: the heat equation component is handled in Fourier space, while the nonlinear term is advanced using a fine forward Euler method. Solutions are computed on a high-resolution grid (\( 8192 \) points) and then subsampled to lower resolutions for training and testing. The goal is to learn the operator mapping \( G^\dagger: L^2_{\text{per}}((0, 1); \mathbb{R}) \to H^r_{\text{per}}((0, 1); \mathbb{R}) \), defined as \( u_0 \mapsto u(\cdot, 1) \) for \( r > 0 \).

\subsubsection{Results and Comparisons}  
Table~\ref{tab:burgers} presents benchmark results for different neural operator methods on the 1D Burgers’ equation across varying resolutions. The proposed Monte Carlo Neural Operator (MCNO) achieves competitive performance, demonstrating its effectiveness in learning operator mappings. Compared to existing approaches, MCNO achieves lower error than GNO, LNO, MGNO, and FNO across all resolutions. Notably, MCNO outperforms FNO on speed and accuracy at every resolution.

The WNO seems to perform best on higher resolutions, the model is not consistent through all resolutions making it not suitable for training on coarse grid and evaluating on finer grid. Moreover, it is slower than FNO and MCNO.

While the MWT Leg model achieves the lowest consistent error, it relies on specialized structured representations and is significantly more computationally expensive. In contrast, MCNO achieves a significant balance between accuracy, efficiency, and adaptability, offering a faster alternative to MWT Leg while maintaining competitive accuracy.

\begin{figure}[h!] \label{L2lossbur}
    \centering
    \includegraphics[width=0.5\textwidth]{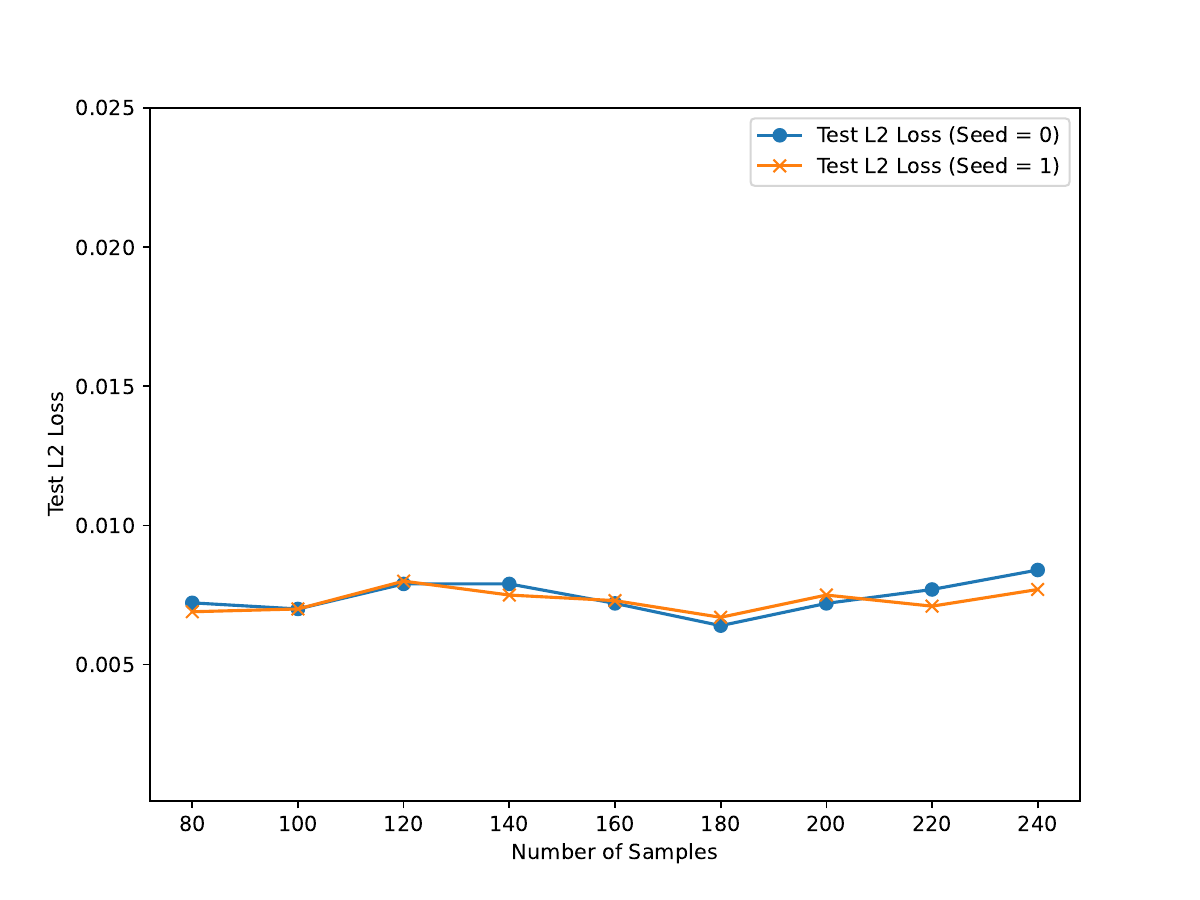} % Adjust width as needed
    \caption{$L_2$ test loss vs Number of samples for Burgers} % Optional: Add a caption
    \label{fig:Picture6} % Optional: Add a label for referencing
\end{figure}

\begin{figure}[h!] \label{timebur}
    \centering
    \includegraphics[width=0.5\textwidth]{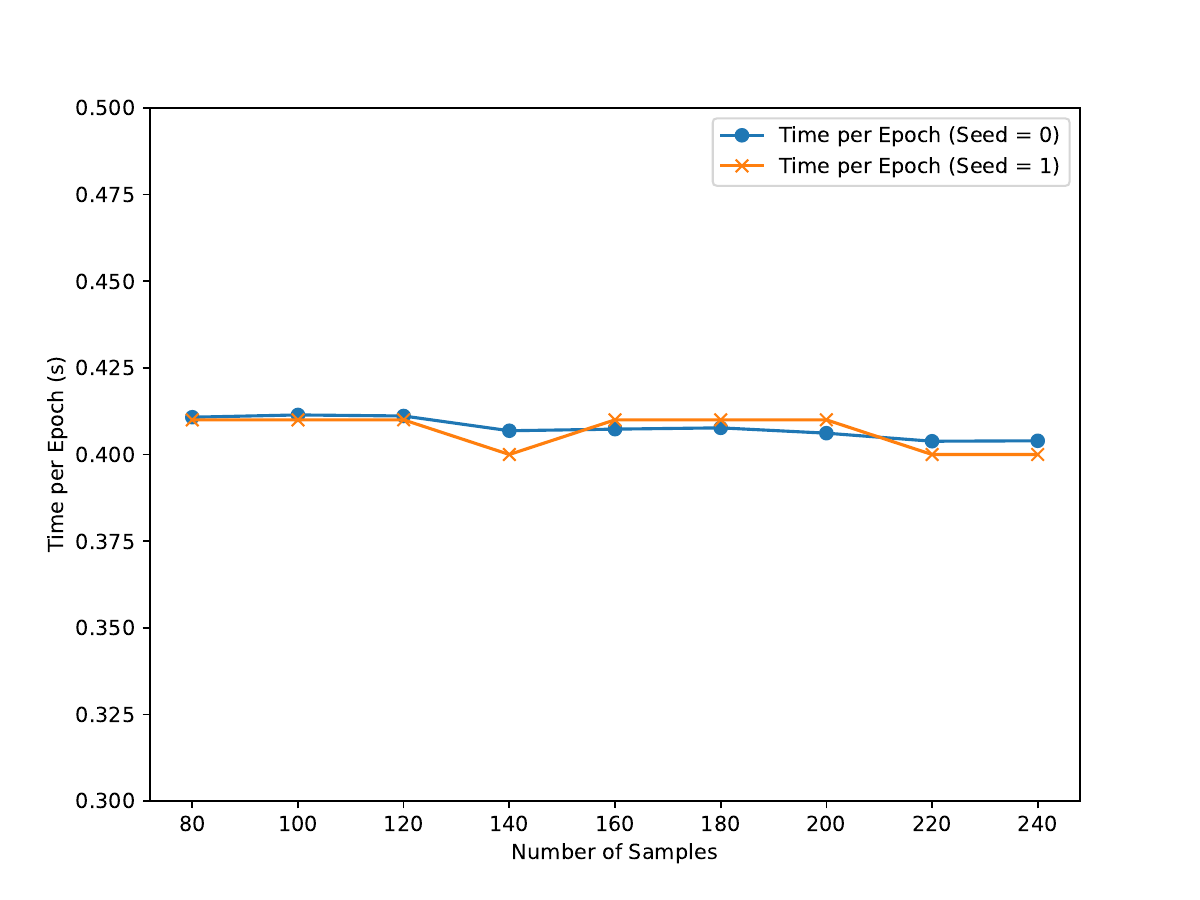} % Adjust width as needed
    \caption{Time per epoch vs Number of samples for Burgers} % Optional: Add a caption
    \label{fig:Picture7} % Optional: Add a label for referencing
\end{figure}

Figure \ref{fig:Picture6} indicates that the relative $L_2$ loss is consistent with the number of samples used. As the number of samples increases the errors tend to be stable. The time per epoch is also consistent with the number of samples showing the efficiency of the MCNO as seen in Figure \ref{fig:Picture7}.

% \begin{table*}[ht]
% \centering
% \caption{Benchmarks on 1-D Burgers’ Equation.}
% \label{tab:burgers}
% \begin{tabular}{|l|c|c|c|c|c|c|}
% \hline
% \textbf{Model} & $s=256$ & $s=512$ & $s=1024$ & $s=2048$ & $s=4096$ & $s=8192$ \\ \hline
% GNO             & 0.0555  & 0.0594  & 0.0651   & 0.0663   & 0.0666   & 0.0699   \\ \hline
% LNO             & 0.0212  & 0.0221  & 0.0217   & 0.0219   & 0.0200   & 0.0189   \\ \hline
% MGNO            & 0.0243  & 0.0355  & 0.0374   & 0.0360   & 0.0364   & 0.0364   \\ \hline
% FNO             & 0.0149  & 0.0158  & 0.0160   & 0.0146   & 0.0142   & 0.0139   \\ \hline
% MCNO            & 0.0076  & 0.0065  & 0.0070   & 0.0071   & 0.0102   & 0.0082    \\ \hline
% MWT Leg          & 0.002  & 0.0019  & 0.0019   & 0.0019   & 0.0019   & 0.0019    \\ \hline
% \end{tabular}
% \end{table*}

\begin{table*}[ht]
\centering
\caption{Benchmarks on 1-D Burgers’ Equation showing relative $L_2$ errors for different input resolutions $s$.}
\label{tab:burgers}
\begin{tabular}{|l|c|c|c|c|c|c|c|}
\hline
\textbf{Model} & \makecell{Time per epoch \\ (s=256, s=8192)} & $s=256$ & $s=512$ & $s=1024$ & $s=2048$ & $s=4096$ & $s=8192$ \\ \hline
GNO            &- & 0.0555  & 0.0594  & 0.0651   & 0.0663   & 0.0666   & 0.0699   \\ \hline
LNO            &- & 0.0212  & 0.0221  & 0.0217   & 0.0219   & 0.0200   & 0.0189   \\ \hline
MGNO           &- & 0.0243  & 0.0355  & 0.0374   & 0.0360   & 0.0364   & 0.0364   \\
\hline
WNO            &  (2.45s, 3.1s) & 0.0546   & 0.0213  &  0.0077  & 0.0043   &  0.0027   &  0.0012  \\ \hline
FNO             & (0.44s, 1.56s) &  0.0183 &  0.0182 &  0.0180  &  0.0177  &  0.0172 & 0.0168   \\ \hline
\textbf{MCNO}      & \textbf{(0.40s, 1.32s)}  & \textbf{0.0064}   &\textbf{0.0067}   & \textbf{0.0062}   & \textbf{0.0069}   & \textbf{0.0071}    & \textbf{0.0065}   \\ \hline
MWT Leg        & (3.70s, 7.10s) & 0.0027  & 0.0026  &  0.0023  &  0.0024  &  0.0025  & 0.0023   \\ \hline
\end{tabular}
\end{table*}

\subsection{Korteweg-de Vries (KdV) Equation}

The Korteweg-de Vries (KdV) equation is a one-dimensional nonlinear partial differential equation used to model shallow water waves and nonlinear dispersive phenomena. Originally introduced by Boussinesq and later rediscovered by Korteweg and de Vries, it is known for its ability to describe solutions and other complex wave interactions. Its dynamics are governed by the equation:
\[
\frac{\partial u}{\partial t} = -0.5 u \frac{\partial u}{\partial x} - \frac{\partial^3 u}{\partial x^3}, \quad x \in (0, 1), \, t \in (0, 1],
\]
where \( u(x, t) \) is the solution field.

\subsubsection{Dataset and Problem Setup.}

The task is to learn the mapping from the initial condition \( u_0(x) = u(x, t=0) \) to the solution at the final time \( u(x, t=1) \). The initial conditions \( u_0(x) \) are sampled from Gaussian random fields with periodic boundary conditions, defined as:
\[
u_0 \sim \mathcal{N}(0, 7^4(-\Delta + 7^2I)^{-2.5}),
\]
where \( \Delta \) represents the Laplacian operator. The equation is numerically solved using the Chebfun package at a high resolution of \( 2^{10} \). Notably, we utilize the same dataset as presented in \cite{gupta2021multiwavelet}, ensuring consistency with prior work. Lower-resolution datasets are generated by systematically sub-sampling the high-resolution solutions.

The nonlinear nature of the KdV equation and its sensitivity to the initial conditions make this a challenging benchmark for operator learning methods. Accurate solutions require capturing both the advection and dispersive behaviors of the equation.

\subsubsection{Results and Comparisons}  
Table~\ref{tab:kdv_results} reports benchmark results for various neural operator models on the Korteweg–de Vries (KdV) equation at different input resolutions. Similar to the Burgers' equation results, MCNO consistently outperforms FNO, MGNO, LNO, and GNO, demonstrating its ability to capture nonlinear dynamics effectively. While MWT Leg achieves the lowest error, MCNO remains a robust alternative, balancing accuracy and computational efficiency through its Monte Carlo-based formulation.

\begin{figure}[h!] \label{L2losskdv}
    \centering
    \includegraphics[width=0.5\textwidth]{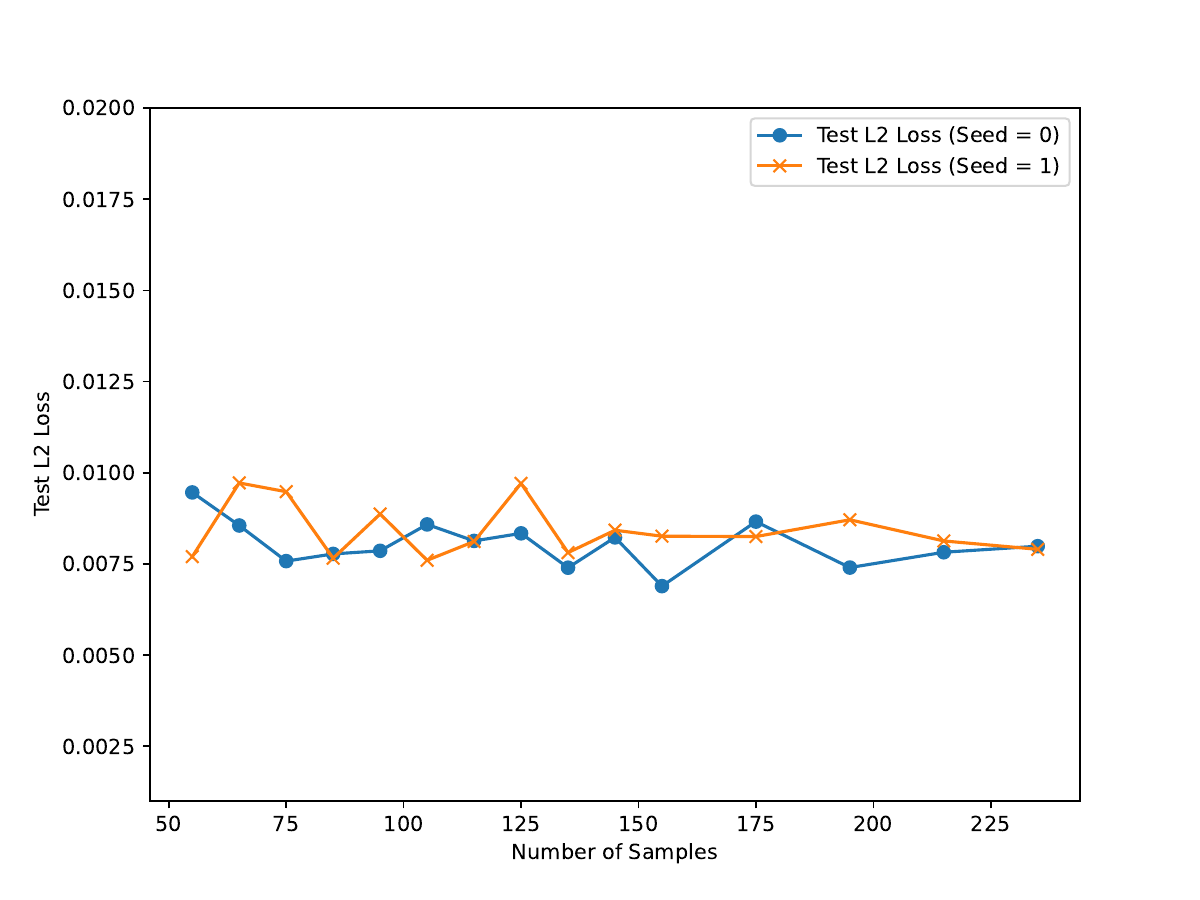} % Adjust width as needed
    \caption{$L_2$ test loss vs Number of samples for KdV} % Optional: Add a caption
    \label{fig:Picture4} % Optional: Add a label for referencing
\end{figure}

\begin{figure}[h!] \label{timekdv}
    \centering
    \includegraphics[width=0.5\textwidth]{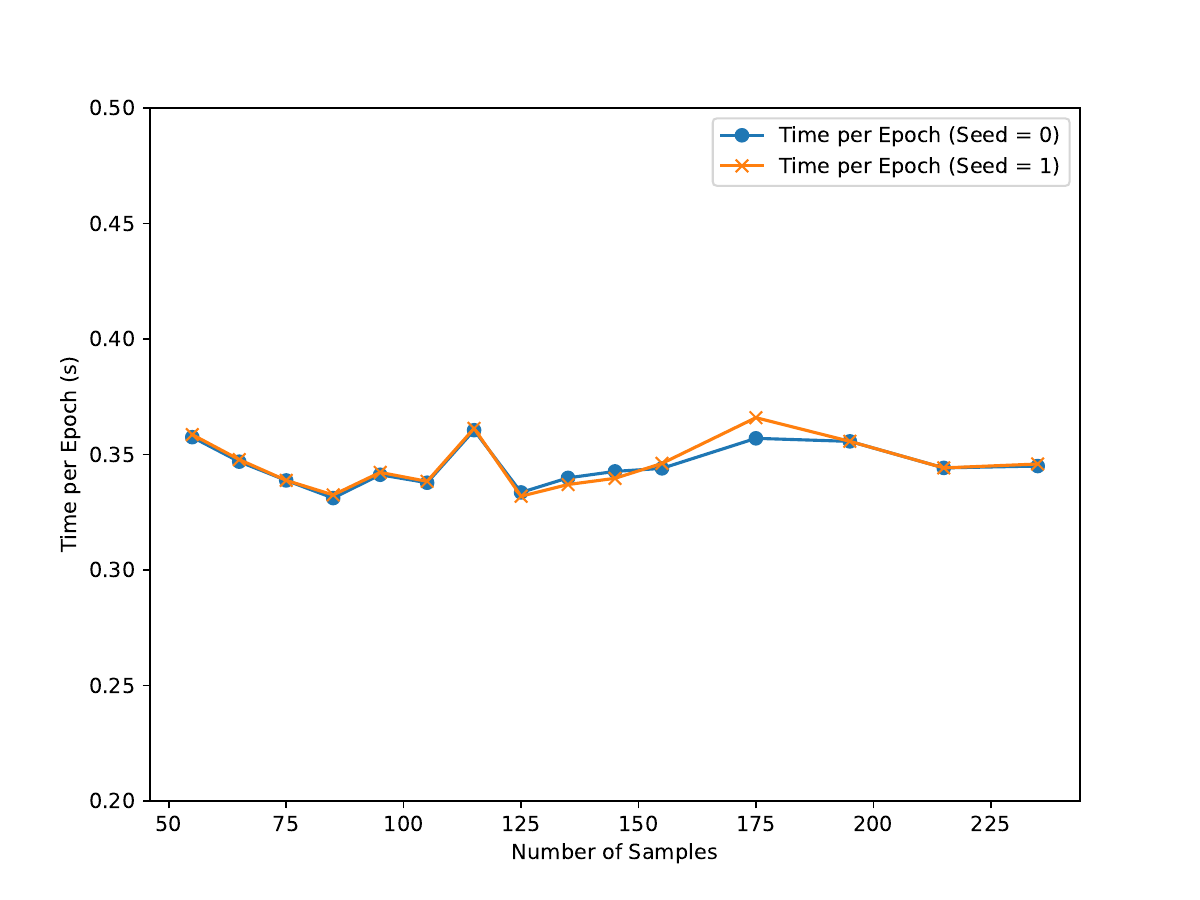} % Adjust width as needed
    \caption{Time per epoch vs Number of samples for KdV} % Optional: Add a caption
    \label{fig:Picture5} % Optional: Add a label for referencing
\end{figure}

Figure \ref{fig:Picture4} shows how the relative $L_2$ loss remains consistent with the number of samples. As the number of samples that are considered increases the results tend to converge. The time taken per epoch does not change much with the number of samples showing the efficiency of the MCNO as seen in Figure \ref{fig:Picture5}.

\begin{table*}[ht]
\centering
\caption{Korteweg-de Vries (KdV) equation benchmarks showing relative $L_2$ errors for different input resolutions $s$.} %Results are reported in terms of relative $L^2$ error. Top: Proposed methods. Bottom: Previous neural operator models.}
\label{tab:kdv_results}
\begin{tabular}{|l|c|c|c|c|c|}
\hline
\textbf{Model}    & \makecell{Time per epoch \\ (s=128, s=1024)}   & \textbf{$s=128$} & \textbf{$s=256$} & \textbf{$s=512$} & \textbf{$s=1024$} \\ \hline
MWT Leg      & (3.24s, 4.96s)    & 0.0036 & 0.0042  & 0.0042  & 0.0040     \\ \hline
\textbf{MCNO} & \textbf{( 0.36s, 0.49s)}     & \textbf{0.0070}  & \textbf{0.0079}   & \textbf{0.0088}   & \textbf{0.0081}     \\ \hline
%MWT Chb      &     & 0.00712 & 0.00604  & 0.00769  & 0.00675     \\ \hline
FNO          &  (0.48s, 0.50s )   & 0.0126  & 0.0125   & 0.0122   & 0.0133     \\ \hline
MGNO         &  -   & 0.1515  & 0.1355   & 0.1345   & 0.1363       \\ \hline
LNO          &  -  & 0.0557  & 0.0414   & 0.0425   & 0.0447       \\ \hline
GNO          &  -  & 0.0760  & 0.0695   & 0.0699   & 0.0721       \\ \hline
\end{tabular}
\end{table*}

\section{Conclusion}

In this work, we introduced the Monte Carlo-type Neural Operator (MCNO), a lightweight operator-learning framework that directly learns kernel functions and employs a single Monte Carlo-type sampling procedure to approximate integral operators. Unlike spectral or hierarchical approaches, MCNO avoids assumptions of translation invariance and does not rely on global basis functions, enabling flexibility across grid resolutions and problem settings.

Through experiments on one-dimensional PDE benchmarks, including Burgers’ and Korteweg–de Vries (KdV) equations, we showed that MCNO achieves competitive accuracy while maintaining computational efficiency and architectural simplicity. Complementing these empirical findings, our theoretical analysis provided guarantees showing that the bias and variance of the Monte Carlo approximation are bounded, with results that extend naturally to higher spatial dimensions. Together, these results position MCNO as an alternative to established neural operator models such as Fourier- and convolution-based operators.

Looking ahead, extending MCNO to higher-dimensional PDEs and more complex systems, such as Navier–Stokes and Darcy flow, represents a natural next step. Further directions include exploring adaptive or problem-aware sampling strategies and evaluating MCNO on unstructured or heterogeneous domains. By combining theoretical guarantees with practical efficiency, MCNO contributes to the growing landscape of neural operators and highlights the promise of Monte Carlo-based approaches for scientific machine learning.

\section*{Acknowledgment}

This work was supported by the NYUAD Center for Interacting Urban Networks (CITIES), funded by Tamkeen under the NYUAD Research Institute Award CG001. The views expressed in this article are those of the authors and do not reflect the opinions of CITIES or their funding agencies

% \section*{Impact Statement}

% This paper aims to advance the field of Machine Learning by introducing novel methodologies and techniques. The potential societal implications of our work are aligned with the broader impacts of advancing computational techniques, including improved efficiency and accuracy in solving complex problems. We have not identified any specific ethical concerns or societal risks directly associated with this research, but we encourage further exploration of the applications of our methods to ensure they are developed responsibly and for the benefit of society.

\bibliographystyle{IEEEtran}

\appendix
% \onecolumn

% \section*{Appendix}
\subsection{\textbf{Proof of Theorem \ref{thm:MCEstimationError}}}
\begin{proof}
We start with the bias-variance decomposition
\begin{multline}
\sup_{x \in D} \left| (\mathcal{K}_\phi v_t)(x) - (\hat{\mathcal{K}}_{N,\phi} v_t)(x) \right| \\
\leq \underbrace{\sup_{x \in D} \left| \mathbb{E}[(\hat{\mathcal{K}}_{N,\phi} v_t)(x)] - (\mathcal{K}_\phi v_t)(x) \right|}_{\text{Bias}} \\
+ \underbrace{\sup_{x \in D} \left| (\hat{\mathcal{K}}_N v_t)(x) - \mathbb{E}[(\hat{\mathcal{K}}_N v_t)(x)] \right|}_{\text{Variance}}.
\end{multline}
% \vspace{2mm}
\noindent \textbf{Analysis of the Bias term:}
% \vspace{2mm}
Since the points $y_{s_i}$ are sampled uniformly from the grid, the expected value of the estimator is the Riemann sum over the entire grid
\begin{equation}
\mathbb{E}\left[ \left( \hat{\mathcal{K}}_{N, \phi} v_t \right)(x) \right] = \frac{1}{N_{\text{grid}}} \sum_{j=1}^{N_{\text{grid}}} \kappa_\phi(x, y_j) v_t(y_j).
\end{equation}
To bound the riemann sum error, we partition $D$ into $N_{\text{grid}}$ hypercubic cells $\{C_j\}_{j=1}^{N_{\text{grid}}}$, each with edge length $h$ and volume $h^d$, centered at $y_j$. The integral over $D$ is
\begin{equation}
\int_D \kappa_\phi(x, y) v_t(y) \, dy = \sum_{j=1}^{N_{\text{grid}}} \int_{C_j} \kappa_\phi(x, y) v_t(y) \, dy.
\end{equation}
Thus, the bias term is
\begin{equation}
\left| \sum_{j=1}^{N_{\text{grid}}} \left( \int_{C_j} \kappa_\phi(x, y) v_t(y) \, dy - \kappa_\phi(x, y_j) v_t(y_j) \cdot h^d \right) \right|.
\end{equation}
The Per-Cell error bound for each cell $C_j$ is
\begin{align}
&\left| \int_{C_j} \kappa_\phi(x, y) v_t(y) \, dy - \kappa_\phi(x, y_j) v_t(y_j) \cdot h^d \right| \\
&\leq \int_{C_j} \left| \kappa_\phi(x, y) v_t(y) - \kappa_\phi(x, y_j) v_t(y_j) \right| \, dy.
\end{align}
By Lipschitz continuity of $\kappa_\phi(x, y) v_t(y)$ for fixed $x$ and since the maximum distance from $y_j$ to any point in $C_j$ is $\frac{\sqrt{d}}{2} h$, thus $\int_{C_j} \| y - y_j \| \, dy \leq \frac{\sqrt{d}}{2} h \cdot h^d$, the per-cell error is
\begin{equation}
\int_{C_j} \left| \kappa_\phi(x, y) v_t(y) - \kappa_\phi(x, y_j) v_t(y_j) \right| \, dy \leq L \cdot \frac{\sqrt{d}}{2} h \cdot h^d.
\end{equation}
Finally we obtain the total bias bound by summing over all cells and recalling that $h = N_{\text{grid}}^{-1/d}$
\begin{align}
\text{Bias}  &= \left| \int_D \kappa_\phi(x, y) v_t(y) \, dy - \frac{1}{N_{\text{grid}}} \sum_{j=1}^{N_{\text{grid}}} \kappa_\phi(x, y_j) v_t(y_j) \right| \\
&\leq \sum_{j=1}^{N_{\text{grid}}} L \cdot \frac{\sqrt{d}}{2} h \cdot h^d =  \frac{\sqrt{d}}{2} \cdot \text{Vol}(D) \cdot L \cdot N_{\text{grid}}^{-1/d}. \label{bia}
\end{align}
% \vspace{5mm}
\noindent \textbf{Bounding the Variance Term:}
% \vspace{5mm}
Define the centered process
\begin{equation}
Z(x) = \left( \hat{\mathcal{K}}_{N, \phi} v_t \right)(x) - \mathbb{E}\left[ \left( \hat{\mathcal{K}}_{N, \phi} v_t \right)(x) \right],
\end{equation}
For any $x, x' \in D$ we have by Lipschitz continuity
\begin{equation}
\left| Z(x) - Z(x') \right| \leq \frac{1}{N} \sum_{i=1}^N L_x \|x - x'\| = L_x \|x - x'\|.
\end{equation}
Thus, $Z(x)$ is Lipschitz continuous with constant $L_x$. To bound the supremum over a continuous domain, construct an $\epsilon$-net $\{x_k\}_{k=1}^M$ of $D$, where for every $x \in D$, there exists an $x_k$ such that $\|x - x_k\| \leq \epsilon$. Since $D \subset \mathbb{R}^d$ is compact, the minimal number of points $M$ in the $\epsilon$-net satisfies:

\begin{equation}
M \leq \left( \frac{2 \cdot \text{diam}(D)}{\epsilon} \right)^d,
\end{equation}
where $\text{diam}(D) = \sup_{x, x' \in D} \|x - x'\|$ is the diameter of $D$. For each fixed $x_k$, $Z(x_k)$ is a sum of $N$ independent random variables, each bounded by $2C$ (since $\left| \kappa_\phi(x_k, y_{s_i}) v_t(y_{s_i}) - \mathbb{E}[\kappa_\phi(x_k, y_s) v_t(y_s)] \right| \leq 2C$). Applying Hoeffding's inequality and taking a union bound
\begin{equation}
\mathbb{P}\left( \max_{k=1,\ldots,M} |Z(x_k)| \geq t \right) \leq 2 M \exp\left( -\frac{N t^2}{2 C^2} \right).
\end{equation}
For any $x \in D$, choose $x_k$ such that $\|x - x_k\| \leq \epsilon$ thus $|Z(x)| \leq |Z(x_k)| + |Z(x) - Z(x_k)| \leq |Z(x_k)| + L_x \epsilon$ and
$\sup_{x \in D} |Z(x)| \leq \max_{k=1,\ldots,M} |Z(x_k)| + L_x \epsilon$. Choosing $\epsilon = \frac{t}{2 L_x}$, we get
$\sup_{x \in D} |Z(x)| \leq \max_{k=1,\ldots,M} |Z(x_k)| + \frac{t}{2}$. Set the probability of exceeding $t$ to be at most $\delta$ with
$2 M \exp\left( -\frac{N t^2}{2 C^2} \right) \leq \delta$, that is 
\begin{equation}
t \geq 2C\sqrt{\frac{2}{N}\left(d\log\left(\frac{4L_x \text{diam}(D)}{t}\right) + \log(2/\delta)\right)}.
\end{equation}
Using $M \leq N_{\text{grid}}$ we get 
\begin{equation}
t \geq 2C\sqrt{\frac{2}{N}\left(d\log N_{\text{grid}} + \log(2/\delta)\right)}.
\end{equation}
Thus, with probability $\geq 1-\delta$
\begin{equation}
\text{Variance} \leq 3C\sqrt{\frac{2}{N}\left(d \log N_{\text{grid}} + \log \frac{2}{\delta}\right)}. \label{var}
\end{equation}
The result is obtained by summing the upper bounds on the bias and variance.
\end{proof}

\subsection{Additional illustrations and Implementations}
\begin{figure}[h!] \label{predkdv}
    \centering
    \includegraphics[width=0.48\textwidth]{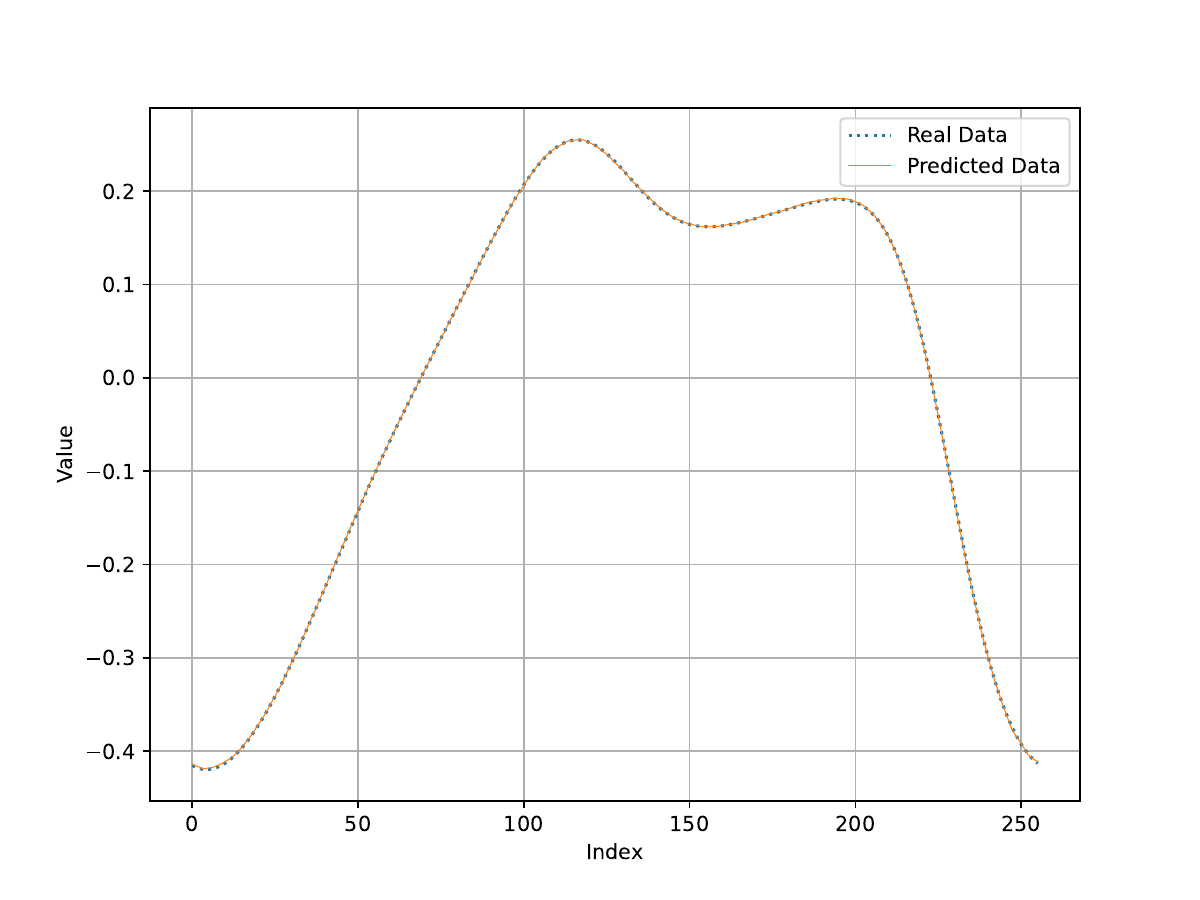} % Adjust width as needed
    \caption{Sample Prediction for Burgers} % Optional: Add a caption
    \label{fig:Picture8} % Optional: Add a label for referencing
\end{figure}

\begin{figure}[h!] \label{predkdv}
    \centering
    \includegraphics[width=0.48\textwidth]{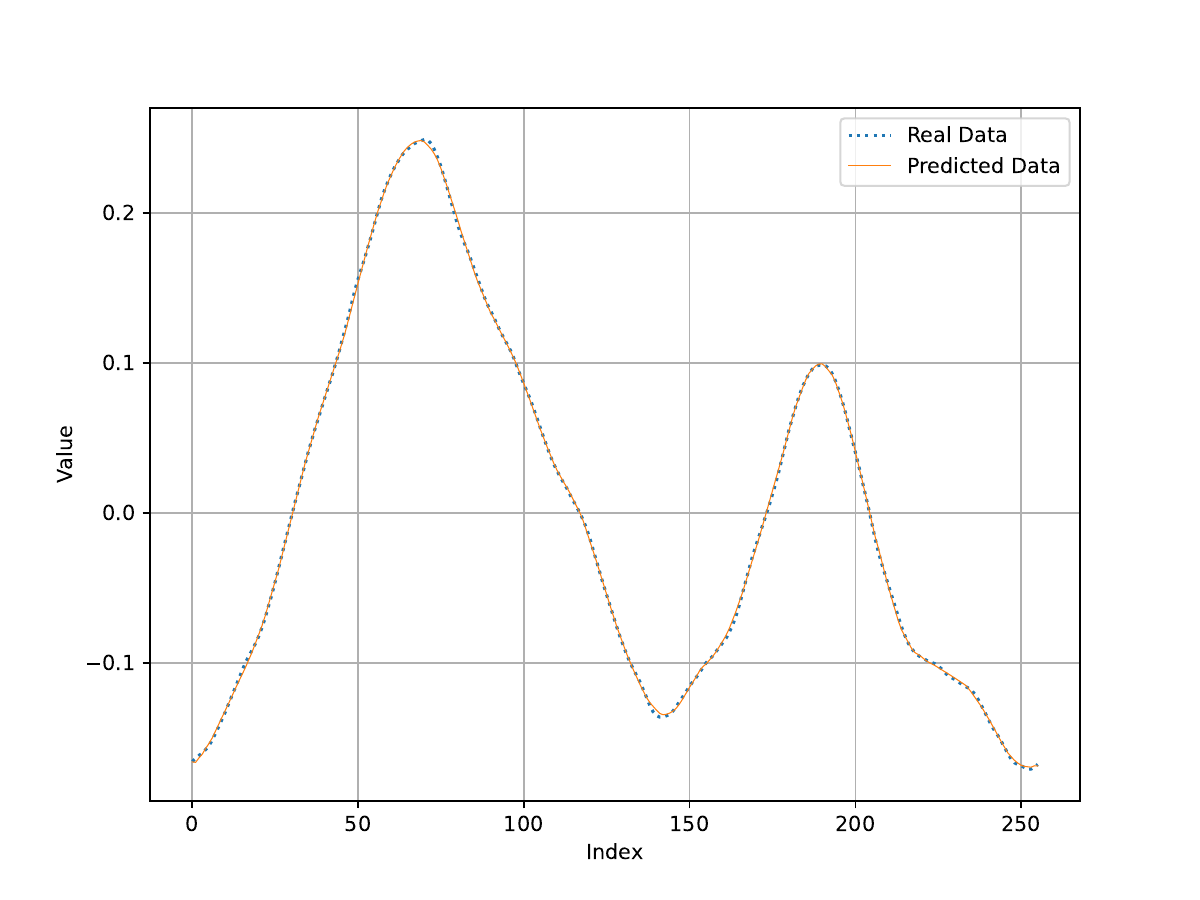} % Adjust width as needed
    \caption{Sample Prediction for KdV} % Optional: Add a caption
    \label{fig:Picture9} % Optional: Add a label for referencing
\end{figure}

\begin{figure}[h!] \label{timekdv12}
    \centering
    \includegraphics[width=0.48\textwidth]{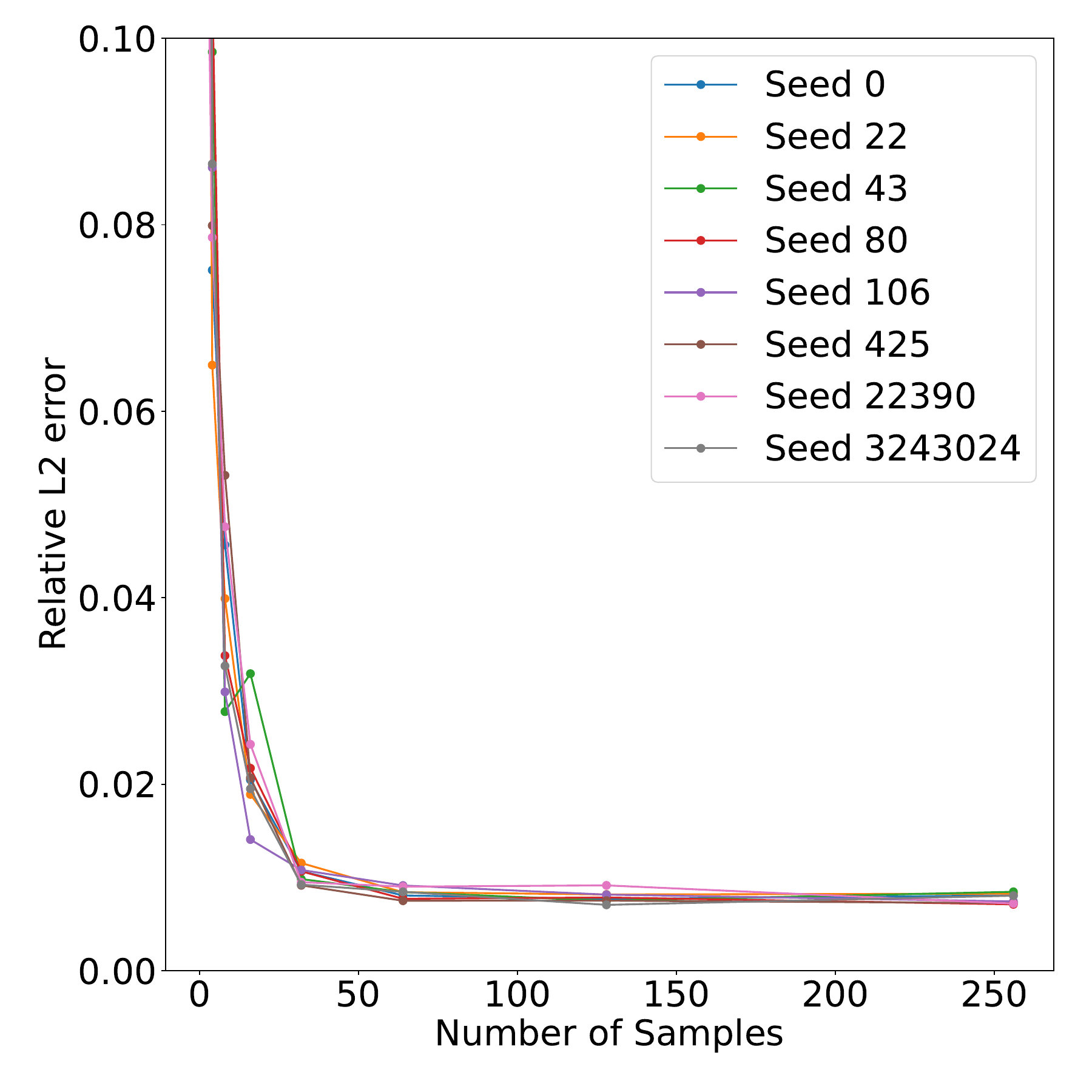} % Adjust width as needed
    \caption{Relative $L_2$ test loss vs Number of samples for KdV} % Optional: Add a caption
    \label{fig:Picture532} % Optional: Add a label for referencing
\end{figure}
\begin{figure}[h!] \label{timekdv213}
    \centering
    \includegraphics[width=0.48\textwidth]{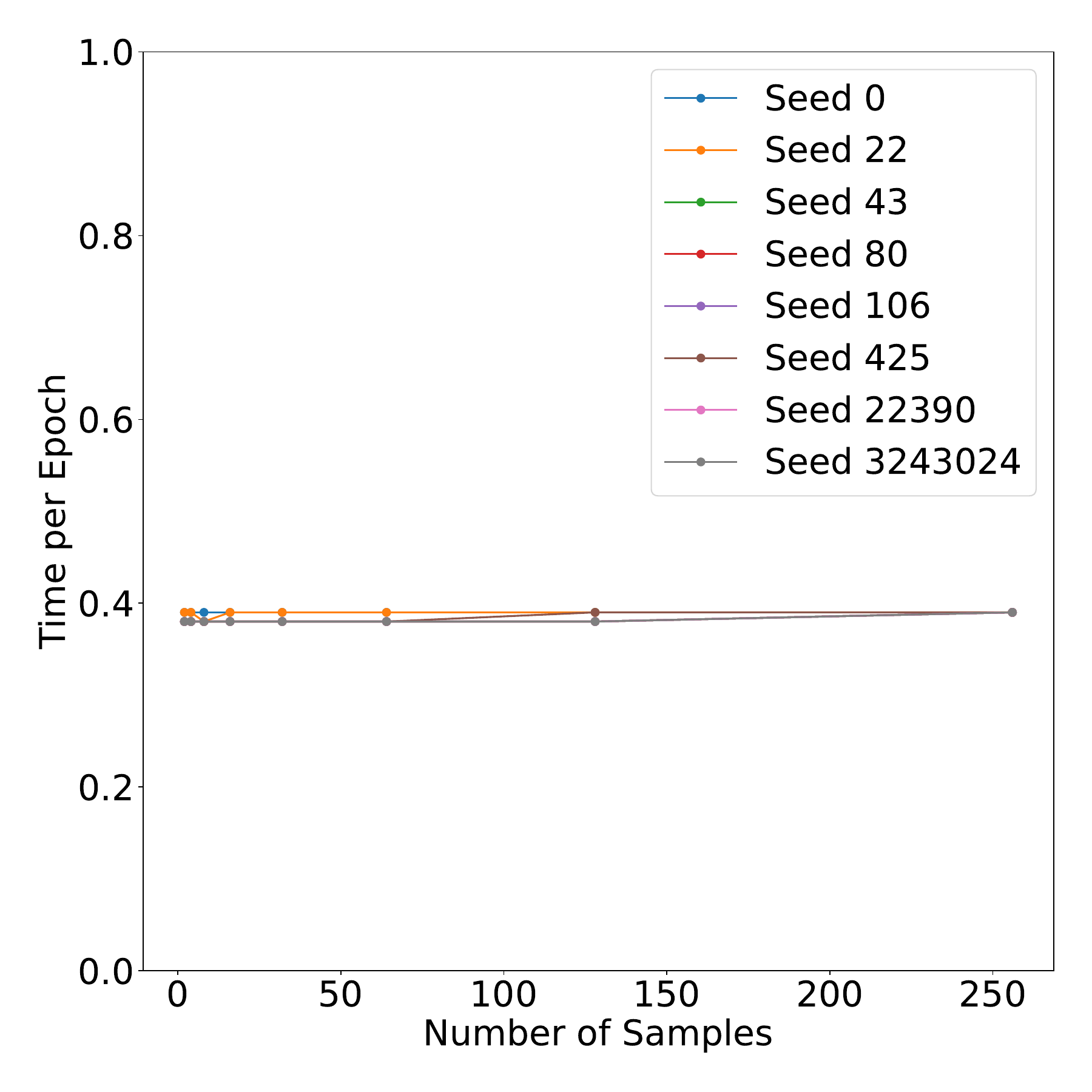} % Adjust width as needed
    \caption{Time per epoch vs Number of samples for KdV} % Optional: Add a caption
    \label{fig:Picture5634} % Optional: Add a label for referencing
\end{figure}

\end{document}